\documentclass{article}

\usepackage[letterpaper]{geometry}
\usepackage[parfill]{parskip}

\PassOptionsToPackage{numbers, compress}{natbib}

\usepackage[numbers, compress]{natbib}
\usepackage{xr-hyper,refcount}

\usepackage{graphicx}
\usepackage[utf8]{inputenc} 
\usepackage[T1]{fontenc}    
\usepackage[colorlinks]{hyperref}       
\usepackage{url}            
\usepackage{booktabs}       
\usepackage{amsfonts}       
\usepackage{nicefrac}       
\usepackage{microtype}      
\usepackage[parfill]{parskip}
\usepackage{algorithm,algorithmic}
\usepackage{amsmath,amsthm,amssymb,bbm}
\usepackage{mathtools}
\usepackage{cases}
\usepackage{comment}
\usepackage{subcaption}
\usepackage{algorithm,algorithmic}
\usepackage{color}
\usepackage{appendix}
\usepackage{xspace}
\usepackage{enumitem}
\usepackage[english]{babel}
\usepackage{authblk}

\usepackage{cleveref}
\crefformat{equation}{(#2#1#3)}
\crefrangeformat{equation}{(#3#1#4) to~(#5#2#6)}
\crefname{equation}{}{}
\Crefname{equation}{}{}

\crefname{definition}{\textbf{definition}}{definitions}
\Crefname{definition}{Definition}{Definitions}
\crefname{assumption}{\textbf{assumption}}{assumptions}
\Crefname{assumption}{Assumption}{Assumptions}
\definecolor{maroon}{RGB}{192,80,77}

\newtheorem{theorem}{Theorem}

\newtheorem{definition}[theorem]{Definition}

	\AtEndDocument{\refstepcounter{theorem}\label{finalthm}}

\usepackage{mdwlist}

\usepackage[utf8]{inputenc} 
\usepackage[T1]{fontenc}    
\usepackage{url}            
\usepackage{booktabs}       
\usepackage{amsfonts}       
\usepackage{amsmath}
\usepackage{nicefrac}       
\usepackage{cleveref}       
\usepackage{natbib}

\usepackage{graphicx}
\usepackage[table]{xcolor}
\definecolor{lightgray}{gray}{0.9}
\usepackage{times}
\usepackage{latexsym}
\usepackage{soul}
\definecolor{hlyellow}{cmyk}{0,0,1,0}
\definecolor{aqua}{rgb}{0.0, 1.0, 1.0}

\usepackage[T1]{fontenc}
\usepackage[utf8]{inputenc}
\usepackage{microtype}
\usepackage{xcolor, soul}
\usepackage{ulem} 
\title{Goodhart's Law Applies to NLP's Explanation Benchmarks}

\author{ 
        Jennifer Hsia$^\dagger$$^*$
        \qquad
        Danish Pruthi$^\ddagger$
        \qquad
        Aarti Singh$^\dagger$
 \qquad
        Zachary C. Lipton$^\dagger$ \\ 
$^\dagger$~Carnegie Mellon University, Pittsburgh, PA \\ 
$^\ddagger$~Indian Institute of Science, Bangalore \\
$^*$\small\texttt{jhsia2@cs.cmu.edu}
}

\begin{document}
\date{}
\maketitle

\begin{abstract}
 Despite the 
rising popularity of 
saliency-based \emph{explanations},
the research community remains at an impasse,
facing doubts concerning their 
purpose, efficacy, and tendency to contradict each other. 
Seeking to unite the community's efforts around
common goals,
several recent works
have proposed evaluation metrics.
In this paper, we critically examine
two sets of metrics:
the ERASER metrics (\emph{comprehensiveness} and \emph{sufficiency})
and the EVAL-X metrics,
focusing our inquiry on natural language processing.
First, we show that we can inflate a model's
comprehensiveness and sufficiency scores dramatically
\emph{without altering its predictions or explanations
on in-distribution test inputs}. 
Our strategy exploits the tendency for extracted \emph{explanations}
and their complements to be ``out-of-support'' 
relative to each other and in-distribution inputs.
Next, we demonstrate that the EVAL-X metrics
can be inflated arbitrarily by a simple method
that encodes the label,
even though EVAL-X is precisely motivated to address such exploits. 
Our results raise doubts about the ability
of current metrics to guide explainability research,
underscoring the need for a broader reassessment
of what precisely these metrics are intended to capture.

\end{abstract}

%
%

\section{Introduction}
\label{sec:intro}

\begin{figure*}[htb]
    \centering
    \includegraphics[width = .99\linewidth]{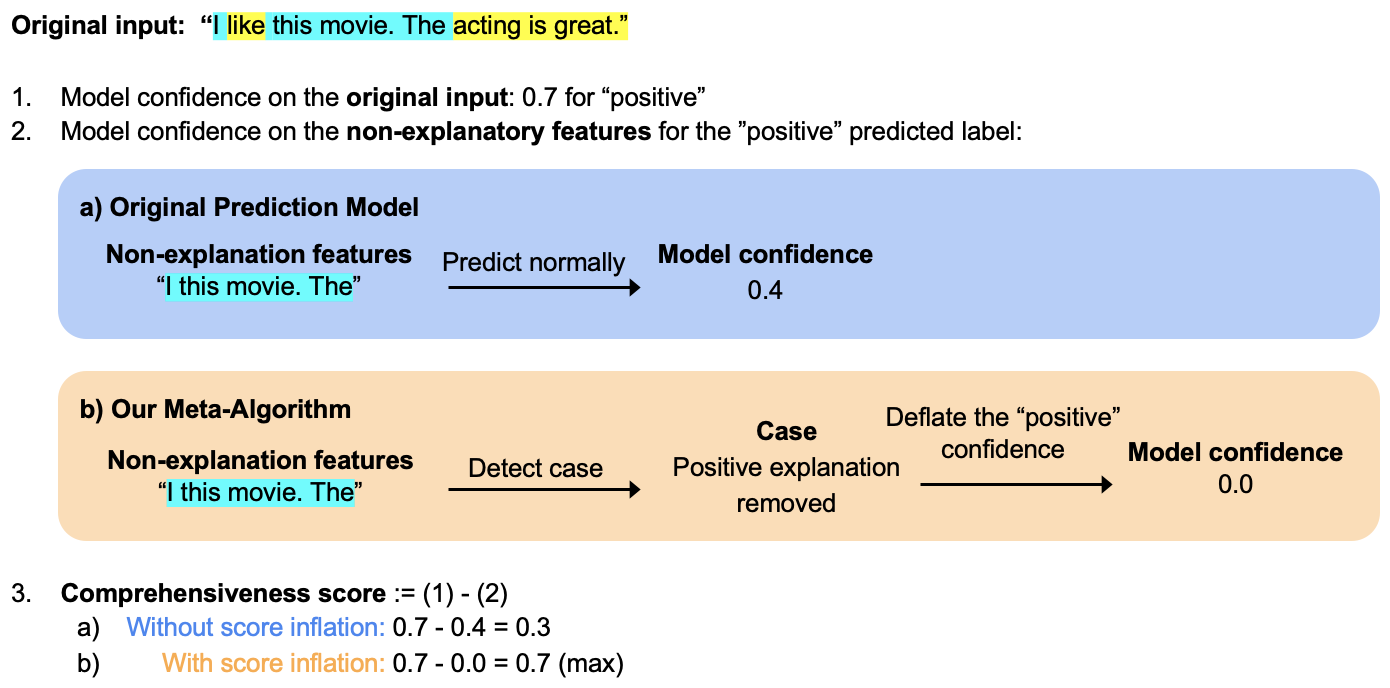}
    \caption{
    The ERASER benchmark's quantitative metrics depend on 
    the prediction model's \emph{confidence} when invoked with
    the original input, \hl{the explanation features}, and \sethlcolor{aqua}\hl{non-explanation features}. 
    In this example movie review, 
    we illustrate how our meta-algorithm can inflate
    comprehensiveness scores without altering 
    predictions or \emph{explanations} on the original inputs.
    Our technique exploits the fact that \emph{explanation-only} features 
    and \emph{non-explanation} features
    are identifiably different.
    }
    \label{fig:motivate}
\end{figure*}

Popular methods for ``explaining'' the outputs 
of \emph{natural language processing} (NLP) models
operate by highlighting a subset of the input tokens
that ought, in some sense, to be \emph{salient}.
The community has initially taken an \emph{ad hoc} approach 
to evaluating these methods, 
looking at select examples
to see if the highlighted tokens align with intuition. 
Unfortunately, this line of research has exhibited
critical shortcomings~\citep{lipton2018mythos}.
Popular methods tend to disagree substantially 
in their highlighted token \emph{explanations} ~\citep{pruthi2022evaluating, krishna2022disagreement}.
Other methods highlight tokens 
that simply encode the predicted label,
rather than offering additional information
that could reasonably be called an \emph{explanation} \citep{jethani2021have}.
This state of affairs has motivated an active area of research
focused on developing evaluation metrics 
to assess the quality of such \emph{explanations},
focusing on such high-level attributes
as \emph{faithfulness}, \emph{plausibility},
and \emph{conciseness}, among others.

In particular, \emph{faithfulness}
has emerged as a primary focus of explainability metrics.
According to \citet{jacovi2020towards}, faithfulness ``refers to how accurately [an explanation] reflects the true reasoning process of the model.''
Such metrics
are typically concerned with how a model's output changes 
when the model is invoked with only the 
\emph{explanatory} tokens 
or when the model receives the non-explanatory tokens
\citep{deyoung2019eraser, agarwal2022openxai, petsiuk2018rise, hooker2019benchmark, serrano2019attention, covert2021explaining,samek2015evaluating,
nguyen2018comparing}. 
Unfortunately, these token subsets 
do not in general resemble natural documents.
Informally, we might say that we are \emph{explaining}
the model by considering its behavior exclusively
on examples that lie outside the support 
of the distribution of interest \citep{hase2021out}.
This raises concerns about whether
degradation in performance 
absent the ostensibly salient features
could be due merely to distribution shift
\citep{hooker2019benchmark}.
This focus on feature subsets in evaluation is no accident;
many \emph{explanation} algorithms determine saliency in the first place
by observing changes in model outputs
on different feature subsets
\citep{ribeiro2016should, lundberg2017unified}.

In this paper, we investigate two sets of \emph{explanation} metrics 
that rely on evaluating the model on masked inputs: 
the ERASER metrics (i.e. comprehensiveness and sufficiency) and the EVAL-X metrics.
We introduce simple algorithms that wrap existing predictors,
achieving near-optimal scores on these faithfulness metrics
\emph{without} doing anything that a reasonable practitioner 
might describe as providing better \emph{explanations}. 
In the ERASER benchmark's case,
we use a simple wrapper model to inflate the 
faithfulness scores
of a given prediction model and saliency method
\emph{while} generating identical predictions and \emph{explanations} on the test set. 
We achieve this by assigning
different model behaviors to the
masked inputs used in faithfulness evaluation
and the original inputs used in prediction and \emph{explanation} generation
(Figure \ref{fig:motivate}).
The second set of metrics, from EVAL-X, 
is advertised as a way to detect when models 
encode predictions in their explanations.
Optimizing for these metrics is claimed to produce 
``high fidelity/accuracy explanations without
 relying on model predictions generated
 by out-of-distribution input''~\citep{jethani2021have}.
Nevertheless, we show that two simple model-agnostic encoding schemes 
can achieve optimal scores, undercutting
the very motivation of the EVAL-X metrics.

While benchmarks seldom capture all desiderata
related to underlying tasks of interest,
significant progress on a well-designed benchmark
should at least result in useful technological progress.
Unfortunately, our results suggest that 
these metrics fail to meet this bar. 
Rather, they exemplify Goodhart's law:
once optimized, they cease to be useful. 
While our results should raise alarms, 
they do not necessarily doom the enterprise
of designing metrics worth optimizing. 
Often, first attempts at technical definitions 
exhibit a speculative flavor,
serving as tentative proposals 
inviting an iterative process 
of community scrutiny and further refinement.
We might look to the development of differential privacy
after years of alternate proposals as inspiration.
That said, our results demonstrate 
considerable challenges that must be overcome
to produce coherent objectives
to guide explainability research.

\section{Related Work}
\label{sec:related}

\noindent\textbf{Evaluating Explanations {}.}
One desideratum of saliency methods is \emph{faithfulness} or \emph{fidelity}, 
described as the ability to capture
the ``reasoning process'' behind a model's predictions~\citep{jacovi2020towards,chan2022comparative}.
\citet{ribeiro2016should} claim that a saliency method is faithful 
if it ``correspond[s] to how the model behaves 
in the vicinity of the instance being predicted''. 
This work has inspired a wave of removal-based metrics 
that measure the faithfulness of a saliency method
by evaluating the model on \emph{neighboring instances},
created by perturbing or removing tokens.
These removal-based metrics can be broadly categorized into:
(i) metrics that assess model behavior 
on the \emph{explanation} features alone; and
(ii) metrics that assess model performance 
on the input features excluding the \emph{explanation} features.
The first category expresses the intuition that ``faithful''
attributions should comprise features \emph{sufficient} 
for the model to make the same prediction with high confidence. 
Our experiments focus on optimizing for a metric called sufficiency \citep{deyoung2019eraser}, 
but other similar metrics include prediction gap 
on unimportant feature perturbation \citep{agarwal2022openxai},
insertion \cite{petsiuk2018rise}, 
and keep-and-retrain \cite{hooker2019benchmark}.
The second category expresses the notion 
that the selected features are \emph{necessary}.
The metric used in our experiments 
is called comprehensiveness \citep{deyoung2019eraser}, 
while many other variations have been proposed, including prediction gap on important feature perturbation \citep{agarwal2022openxai}, 
deletion \cite{petsiuk2018rise}, remove and retrain \citep{hooker2019benchmark}, 
JS divergence of model output distributions \citep{serrano2019attention},
area over perturbation curve \citep{samek2015evaluating}, 
and switching point \citep{nguyen2018comparing}.
Notably, \citet{jethani2021have} are 
less concerned with ``explaining the model''
and more concerned with justifying the label;
their evaluation checks the behavior of, EVAL-X,
an independent evaluator model 
(not the original predictor),
when invoked on the \emph{explanation} text.

\noindent\textbf{The ``Out-of-Support'' Issue {}.}
One issue has emerged to reveal critical shortcomings 
in these current approaches to saliency:
they attempt to ``explain'' a model's behavior
on some population of interest (e.g., natural documents)
by evaluating how the model behaves on
a wildly different population (the documents that result 
from masking or perturbing the original documents)~\citep{hooker2019benchmark, slack2020fooling}.
Among proposed patches, \citet{hooker2019benchmark} 
create modified training and test sets
by removing the most important features 
according to their attribution scores,
then retraining and evaluating the given model on the modified datasets. 
While such patches address a glaring flaw,
we still lack an affirmative argument for their usefulness;
the OOD issue reveals a fundamental problem
that does not necessarily resolve when the OOD issue is patched.
Moreover, the retrained model is no longer
the object of interest that we sought to explain in the first place.
Others have tried to bridge the distribution gap 
by modifying only the training distribution. 
\citet{hase2021out} suggest modifying the training set 
by adding randomly masked versions of each training instance,
thus all masked inputs would technically be in-distribution.
Although \citet{hase2021out} mention that it is possible
to game metrics when the masked samples are out-of-distribution, 
they do not demonstrate this.
We offer concrete methods to demonstrate not only 
\textit{how easy} it is to optimize removal-based faithfulness metrics, 
but also \textit{how much} these metrics can be optimized.
Following a related idea, \citet{jethani2021have} 
introduce an evaluator model EVAL-X
that is trained on randomly masked inputs from the training data.
Their metrics consist of the EVAL-X's accuracy and AUROC 
when invoked on \emph{explanation-only} inputs.
While the authors claim that EVAL-X can distinguish 
between extract-then-classify models
that encode and those that do not,
we demonstrate two encoding methods
that are scored optimally by EVAL-X,
revealing a critical shortcoming.

\noindent\textbf{Manipulating Explanations {}.}
\citet{slack2020fooling} demonstrate how one could exploit the OOD issue 
to manipulate the feature importance ranking from LIME and SHAP
and conceal problems vis-a-vis fairness.
They propose an adversarial wrapper classifier 
designed such that a sensitive feature 
that the model truly relies on
will not be detected as the top feature.
\citet{pruthi19learning} demonstrate the manipulability
of attention-based \emph{explanations} and
\citet{wang2020gradient} the manipulability of gradient-based \emph{explanations} in the NLP domain.
Many have also explored the manipulability of saliency methods but in the image domain
\citep{heo2019fooling, dombrowski2019explanations, ghorbani2019interpretation}.
In a more theoretical work, 
\citet{anders2020fairwashing} use differential geometry
to establish the manipulability of popular saliency methods.
\textbf{Key difference:} while these works are concerned
with manipulating the \emph{explanations} themselves,
we are concerned with manipulating the leaderboard.

\section{Optimizing the ERASER Benchmark Metrics}
\label{sec:eraser}
\begin{figure*}[htb]
    \centering
    \includegraphics[width = .99\linewidth]{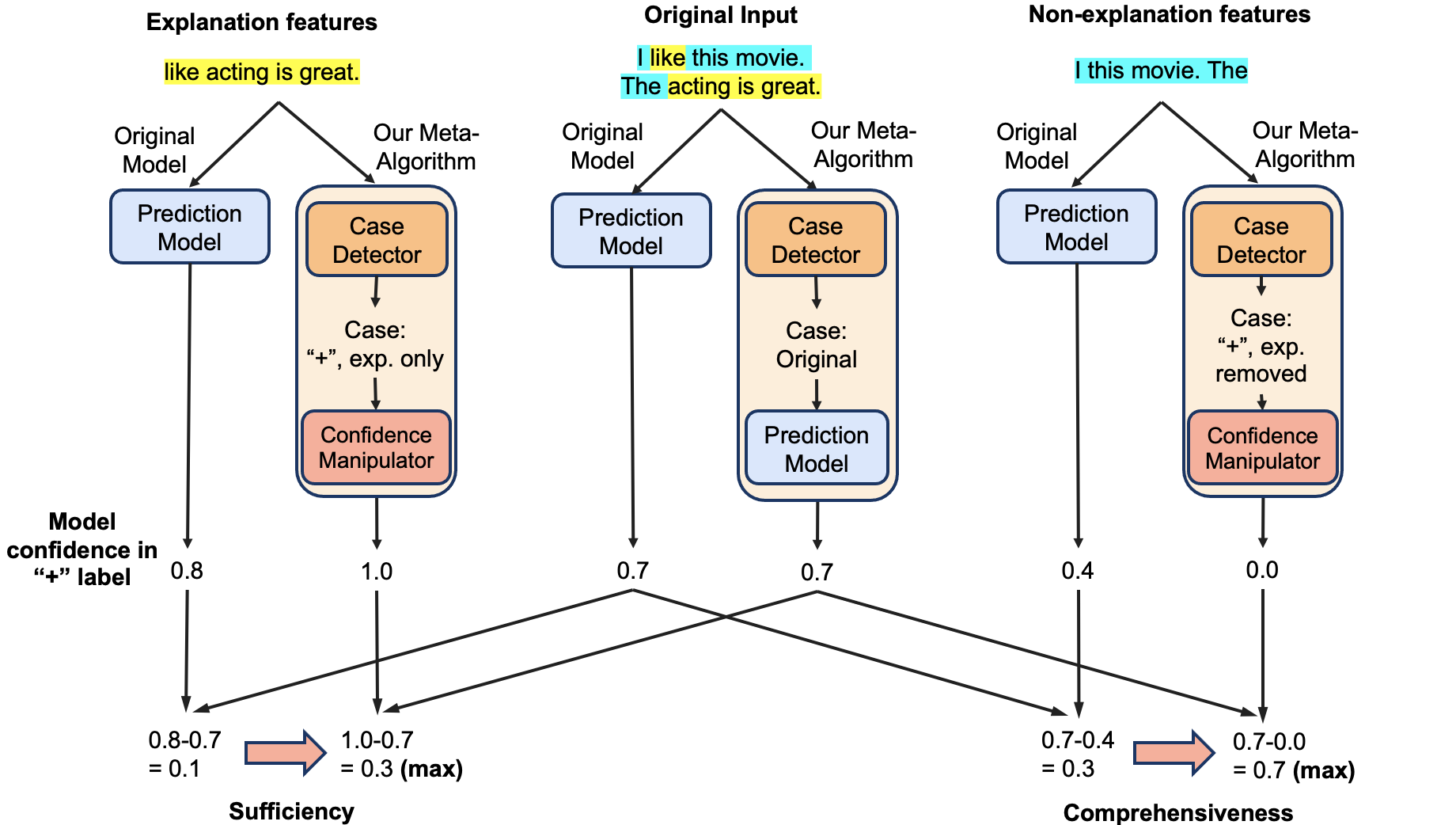}
    \caption{Our meta-algorithm, which wraps a prediction model and saliency method,
    applied to a movie review in a sentiment analysis task.
    First, our case detector determines whether the input 
    consists of (Left) \sethlcolor{hlyellow}\hl{the \emph{explanation-only} features 
    for a particular predicted label (left)},
    (Middle) an original input $x$ (middle),
    or (Right) \sethlcolor{aqua}\hl{the \emph{non-explanation} features 
    for a particular label (right)}. 
    Then if the case is original, 
    we return the probabilities output by the original prediction model.
    Otherwise, our meta-algorithm manipulates the model confidence
    to inflate the sufficiency and comprehensiveness scores.
    }
    \label{fig:eraser}
\end{figure*}

Let $x$ denote a sequence of input tokens,
$y \in \{1, \dots, |\mathcal Y |\}$ a categorical target variable,
and $f$ a prediction model
that maps each input to 
a predicted probability over the $|\mathcal Y |$ labels.
By $\hat y$, we denote the predicted label, 
and $\hat{e}$ a generated \emph{explanation}
consisting of an ordered subset of the tokens in $x$.
By $x \setminus \hat{e}$, we denote the \emph{non-explanation} features
that result from deleting the \emph{explanation}.

\begin{definition}[Sufficiency]
Sufficiency is the difference between the model confidence (on the predicted label) given only the \emph{explanation} features
and the model confidence given the original input:
\begin{equation}
  f(Y = \hat y|X = \hat e) - f(Y = \hat y|X = x).
  \end{equation}
\end{definition} 
Note that our definition is a negation 
of the original sufficiency metric \citep{deyoung2019eraser}.
We make this change for notational convenience 
and to reflect the intuition that
sufficiency is a positive attribute:
higher sufficiency should be better.

\begin{definition}[Comprehensiveness]
Comprehensiveness
is the difference 
between the model confidence 
given the \emph{non-explanation} features 
and the model confidence given the original input:
\begin{equation}
    f(Y = \hat y |X = x) - f(Y = \hat y|X = x\setminus \hat e).
\end{equation}
\end{definition}
Intuitively, a higher comprehensiveness score is thought to be better 
because it suggests the \emph{explanation} captures
most of the ``salient'' features, 
making it difficult to predict accurately in its absence.

For a given prediction model and saliency method, 
we aim to increase the sufficiency and comprehensiveness 
scores while preserving the original predictions and \emph{explanations}.
Let the model 
confidence in the original inputs be $f(Y = \hat y|X = x) = c$.
Then, sufficiency has a range of $[-c, 1-c]$, and is maximized when we set
$f(Y = \hat y|X = \hat e)$ to $1$.
Comprehensiveness has a range of $[c - 1, c]$, and is maximized when 
we set 
$f(Y = \hat y|X = x \setminus \hat e)$ to $0$.
However, there is a tradeoff between these two metrics, where
the two metrics depend on $c$
in opposite directions. 
To maximize sufficiency, we must minimize $c$, 
for which the lowest possible value approaches $1/|\mathcal Y |$ 
(any lower and we change the predicted class). 
On the other hand, to maximize comprehensiveness, we must maximize $c$.
The upshot of this tradeoff is that the sum of sufficiency
and comprehensiveness scores 
lies in the range $[-1, 1]$
and thus cannot exceed $1$.

\subsection{Method}
The key to our method for maximizing these scores
begins with the insight that 
\emph{explanation-only} inputs $\hat e$
and \emph{non-explanation} inputs $x \setminus \hat e$ 
are easy to distinguish from original inputs $x$.
Thus, by recognizing which case we face,
our model can output strategically chosen confidence scores
that inflate the resulting faithfulness scores.
To instantiate this idea, we implement a case detector,
trained to recognize whether an input 
is (i) an original input $x$;
(ii) the \emph{explanation-only} features for a particular label;
or (iii) the \emph{explanation-removed} features for a particular label. 
As a result, our case detector must choose among
$2|\mathcal Y| + 1$ cases
where $|\mathcal Y|$ 
is the number of classes.
For any (prediction model, saliency method) pair,
we must train a fresh case predictor..
Given such a pair, we construct a training set
that consists of every instance in the original train set,
the \emph{explanation-only} features for that instance,
and the \emph{non-explanation} features for that instance. 
The corresponding labels are produced straightforwardly,
e.g., ``an explanation-only input whose predicted label was class $j$''.

Our \textbf{meta-algorithm} wraps the original predictor 
as follows (Figure~\ref{fig:eraser}): 
if the detected case is original, 
we run the input through the original model, 
thereby preserving the same prediction $\hat y$ and \emph{explanation} $\hat{e}$.
If the detected case is \emph{explanation} features for label $y$,
we manually set the model confidence to 1 for label $y$, and 0 for the other labels.
If the detected case is \emph{explanation-removed} features for a label $y$,
we set the model confidence to 0 for label $y$, 
and 1 for a random other label $\neq y$.
If the case predictor is perfectly accurate, 
this procedure achieves a sufficiency score of $1-c$
and the comprehensiveness score $c$,
reaching Pareto optimality.

\subsection{Experimental Setup}
We assess the efficacy of our meta-algorithm
for inflating the sufficiency and comprehensiveness metrics 
using the same datasets as in the original ERASER benchmark paper \citep{deyoung2019eraser}.
We present the results for the Movies \citep{zaidan2008modeling} 
and BoolQ \citep{clark2019boolq} datasets in the main paper 
and share the remaining results for other datasets including
Evidence Inference \citep{lehman2019inferring}, 
FEVER \citep{thorne2018fever},
and MultiRC \citep{khashabi2018looking} 
in the Appendix (Tables \ref{tab:eraser} and \ref{tab:eraser_cd}).

We use pre-trained BERT tokenizers and models~\cite{devlin2018bert} 
for the case detectors and the prediction models. 
We train the prediction models for $10$ epochs 
and the case detector models for $3$ epochs,
both with a batch size of $32$,
and a learning rate of $2$e$-5$. 
We experiment with several saliency methods,
including LIME \citep{ribeiro2016should},
Integrated Gradients (IG) \citep{sundararajan2017axiomatic},
Attention \citep{xu2015show}, 
and a random baseline (which randomly highlights tokens). 
For each saliency method, we use the top 10\% of the input features 
with the highest attribution scores as the \emph{explanation}.
We train a different case detector 
for each prediction model and saliency method pair.
We use a macro-averaged F1 score for the prediction model's task performance
and comprehensiveness and sufficiency for faithfulness.

\subsection{Results}
    
\begin{table*}[tb]
    \centering
\scalebox{0.95}{
\begin{tabular}{l|cccc|cccc}
\toprule

\multicolumn{1}{c}{{}} & \multicolumn{4}{c}{\textbf{Movies}} & \multicolumn{4}{c}{\textbf{BoolQ}}\\
\hline
\textbf{Method}  & \textbf{F1 Score} & \textbf{Comp.} & \textbf{Suff.} & \textbf{Comp.+Suff.} & \textbf{F1 Score} & \textbf{Comp.} & \textbf{Suff.} & \textbf{Comp.+Suff.}  \\ 
\hline
Attention              & 92.4  & 0.18   & -0.11  & 0.07 & 58.4              & 0.05           & -0.01           & 0.04      \\
\rowcolor[HTML]{E7E6E6} 
+ meta-algo & 92.4              & \textbf{0.89}            & \textbf{-0.09}            & \textbf{0.80}    &58.4              & \textbf{0.59}           & \textbf{0.16}            & \textbf{0.75}             \\ \hline
IG               & 92.4              & 0.26 & \textbf{-0.08}     & 0.18   & 58.4              & 0.03           & 0.00            & 0.04  \\
\rowcolor[HTML]{E7E6E6} 
+ meta-algo  & 92.4              & \textbf{0.83}            & -0.09          & \textbf{0.74}   & 58.4              & \textbf{0.73}           & \textbf{0.25}            & \textbf{0.98}               \\ \hline
LIME                   & 92.4              & 0.38      & -0.01   & 0.37  & 58.4              & 0.09           & 0.08            & 0.16  \\
\rowcolor[HTML]{E7E6E6} 
+ meta-algo      & 92.4              & \textbf{0.82}  & \textbf{0.00}           & \textbf{0.82}     & 58.4              & \textbf{0.73}           & \textbf{0.26}            & \textbf{1.00}                \\ \hline
Random                 & 92.4              & 0.01   & -0.06 &-0.05   & 58.4              & 0.01           & -0.06           & -0.05  \\
\rowcolor[HTML]{E7E6E6} 
+ meta-algo    & 92.4              & \textbf{0.65} & \textbf{0.12}   & \textbf{0.77} & 58.4              & \textbf{0.65}           & \textbf{0.12}            & \textbf{0.77}    \\    
\bottomrule

\end{tabular}}
    \caption{
    We demonstrate the comprehensiveness (comp.) 
    and sufficiency (suff.) gains of our meta-algorithm 
    on the ERASER Benchmark's Movies and BoolQ datasets.
    We maintain the same predictions on the original inputs, 
    hence there are no changes in the F1 score. 
    At the same time, on the Movies dataset,
    we achieve a $0.59$ gain in comprehensiveness, 
    and $0.05$ gain in sufficiency, 
    when averaged across these model-saliency method pairs. 
    On the BoolQ dataset,
    we achieve a $0.63$ average comprehensiveness gain 
    and $0.20$ average sufficiency gain.
    }
    \label{tab:eraser_movies}
\end{table*}

\begin{figure*}[htb]
    \centering
    \includegraphics[width = .49\linewidth]{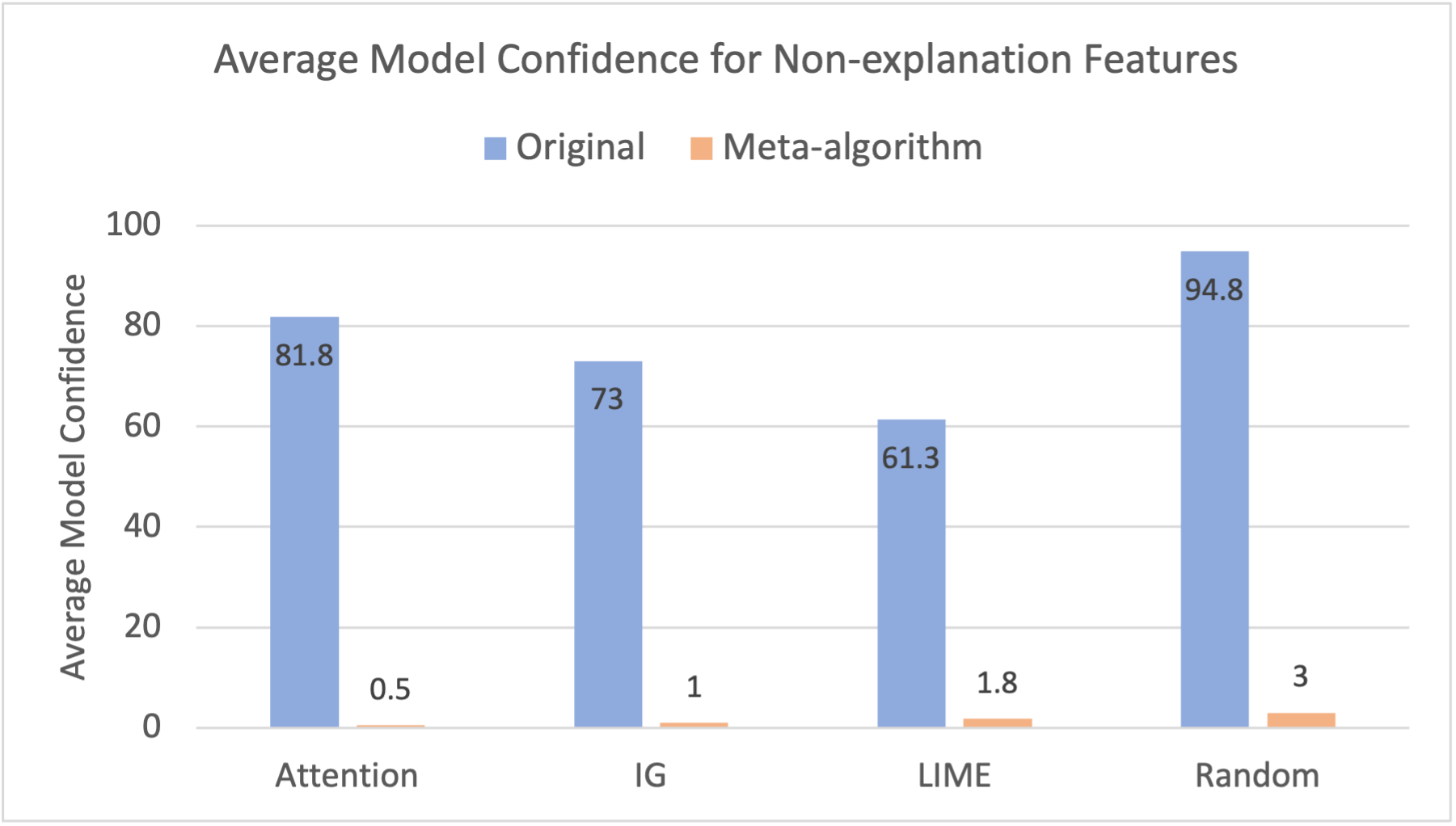}%
    \includegraphics[width = .49\linewidth]{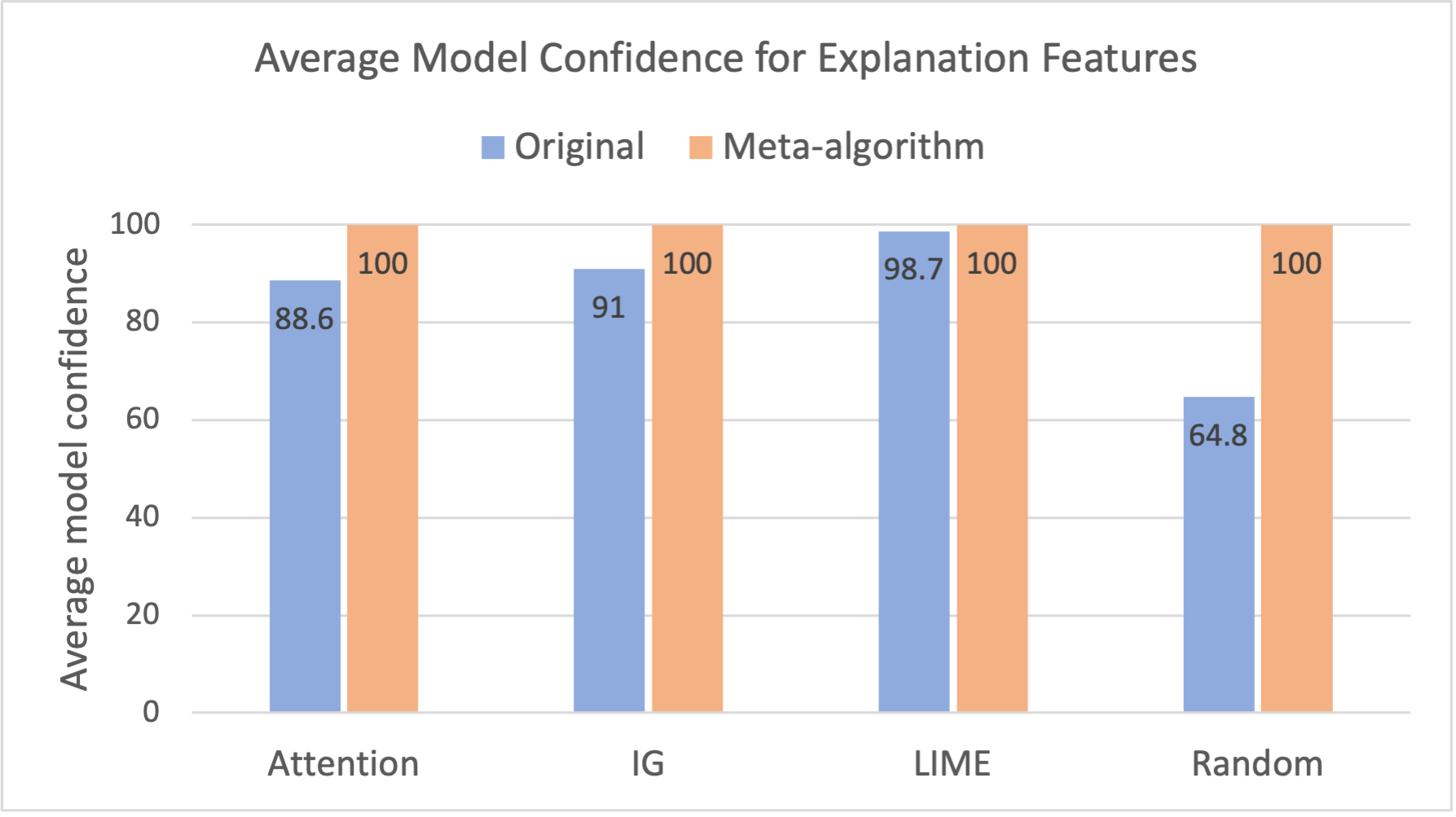}
    \caption{We compare the model confidence in \emph{explanation} and \emph{non-explanation} features from the original model and our meta-algorithm on the Movies dataset. (Left): The optimal comprehensiveness is achieved when the model confidence in \emph{non-explanation} features is 0\%.
    Since the original confidence in \emph{non-explanation} features is high (77.7\% on average), there is a large room to deflate the confidence for comprehensiveness gain.
    In practice, our meta-algorithm method achieves $<5\%$ average confidence, which is close to optimal.
    (Right): The optimal sufficiency is achieved when the model confidence in \emph{non-explanation} features is 100\%. 
    Since the original model's confidence in \emph{explanation} features 
    is already high (85.8\% on average), there is little room to inflate it for sufficiency gain.
    In practice, our meta-algorithm achieves 100\% confidence.
    }
    \label{fig:confidence}
\end{figure*}

Across all the investigated setups, 
our meta-algorithm
is effective in increasing the 
comprehensiveness and sufficiency scores.
For instance,
on the Movies dataset,
with attention-based \emph{explanations}
the initial comprehensiveness score 
was $0.18$, 
but we inflate it to $0.89$ (Table \ref{tab:eraser_movies}).
Similarly, on the BoolQ dataset,
for the IG method,
we again see a dramatic increase, from $0.03$ to $0.73$.
On average, on the Movies dataset,
our meta-algorithm has a comprehensiveness gain of $0.59$
and a sufficiency gain of $0.05$.
Similarly, on the BoolQ dataset,
our meta-algorithm's average comprehensiveness gain is $0.63$ and sufficiency gain is $0.20$.
To put these gains in perspective, recall that the sum
of comprehensiveness and sufficiency cannot exceed $1$.

As one may note, 
the comprehensiveness gains are larger than the sufficiency gains.
This is because the headroom for comprehensiveness gains 
exceeds that for sufficiency gain in practice.
The comprehensiveness gains are bounded 
by how close the original confidence scores are to 0\%
for \emph{non-explanation} features.
In practice, on the Movies dataset,
we observe that the original confidence 
for \emph{non-explanation} features is 77.7\%
(far from 0\%), 
indicating a large potential for score improvement (Fig. \ref{fig:confidence}).
On the other hand, the 
room for inflating sufficiency 
is capped by 
how close the original confidence scores 
for \emph{explanation} features are to 100\%.
For the Movies dataset,
the original model confidence for \emph{explanation} features is 85.8\%  (close to 100\%), 
indicating a smaller potential for score improvement (Fig. \ref{fig:confidence}).

Using our meta-algorithm, 
we minimize the average model confidence 
for \emph{non-explanation} features 
to 1.6\% (close to the optimal 0\%)
and maximize the confidence for \emph{explanation} features 
to the optimal 100\%.
We also compare the sum of the comprehensiveness 
and sufficiency scores in the last column of 
Table \ref{tab:eraser}.
For any given prediction model and saliency method pair, 
our meta-algorithm shows substantial gains in faithfulness sum score.
On average, on the Movies dataset, our meta-algorithm's
sum faithfulness score is 0.78,
whereas the underlying method's faithfulness sum score is 0.14.
On BoolQ, our meta-algorithm's faithfulness sum score is 0.88
whereas the underlying method's score is 0.05.
In some instances, we even achieve the exact optimal score of 1, 
as seen when our meta-algorithm is applied with LIME for BoolQ.
The main reason why our scores are not always 1
is that our case detector 
does not always have perfect test accuracy (Table~\ref{tab:eraser_cd}).

If one took these scores at face value,
our improved faithfulness scores would appear to suggest 
that the \emph{explanations} from our meta-algorithm 
are substantially more faithful
than the \emph{explanations} from the original, non-optimized methods.
However, on original (not masked) inputs,
we typically output the same predictions 
and \emph{explanations} 
as the original models. 
Our ability to max out these benchmarks 
without even changing the explanations themselves
(on the population of interest)
suggest that these metrics are not suited
to guide advances in explainability research. 

Another alarming observation is that 
our optimized version of \textbf{random \emph{explanations}}
has higher faithfulness scores than the non-optimized version 
of the other saliency methods.
A random \emph{explanation} is generated 
without interaction with the prediction model, 
so one would typically expect it to be less faithful 
than other proposed saliency methods.
However, using our meta-algorithm,
the random \emph{explanations} achieve higher faithfulness scores,
raising further doubts about the reasonableness of these scores.

\section{Optimizing scores on EVAL-X Metrics}

The EVAL-X metrics are focused on 
the extract-then-classify variety of ``explainable'' classifiers. 
They confront the issue that when 
an \emph{explanation} extractor and label predictor 
are trained jointly,
the extractor may end up doing all of the work
by simply ``encoding" the eventual prediction,
rather than providing evidence. 
Consider for instance, 
on a binary classification task, 
an \emph{explanation} extractor 
that outputs a period whenever the prediction is positive,
and a comma whenever the prediction is negative. 
The classifier can perfectly recover
the predicted label from the single token, 
encoded \emph{explanation}.
This issue has been highlighted in several past works, 
including~\cite{treviso2020explanation} and \cite{pruthi2022evaluating}.
To detect such encoded \emph{explanations},
\citet{jethani2021have} propose evaluating the \emph{explanations}
under the ``true'' posterior of the predicted label.
Concretely, what they propose
is to train an \emph{evaluator model} called EVAL-X.
During training, EVAL-X is given training 
examples consisting of randomly masked inputs
together with the true corresponding target.
The proposed evaluation strategy for any explanation method
is to assess the accuracy (\textbf{eACC})
and AUC (\textbf{eAUROC}) 
of the EVAL-X model on the masked input
consisting only of \emph{explanation} tokens.
\citet{jethani2021have} suggest that encoding 
in extract-then-classify models
should be detectable as a sharp decline 
in eACC and eAUROC compared to ACC and AUROC.
We define a sharp decline as a 10\% drop.

\subsection{Method}
To game this metric, we propose two simple encoding methods
that achieve high eACC and eAUROC (close to ACC and AUROC).
Formally, if the \emph{explanation} is generated using only the predicted label 
and no other access to the prediction model, 
then the \emph{explanation} encodes the prediction.
Given such an \emph{explanation}, 
the extract-then-classify model can then 
confidently map the \emph{explanation} to the encoded label 
by accessing a fixed code book.

\paragraph{Method 1: Token-label Likelihood Ratio} 
For each label $Y$ and token $T$ 
in the training set,
we calculate the frequency with which the token $T$
appears among instances associated with the label $Y$.
At test time, for a given input,
we first use the prediction model 
to get $\hat y$.
Then, to generate an encoded \emph{explanation},
we output the $k$ input tokens
that have the highest likelihood ratio
(token frequency under label $\hat y$ 
divided by token frequency under labels $Y \neq \hat{y}$).

\paragraph{Method 2: Evaluator Model Queries}
Alternatively, if we know the exact weights of the evaluator model, 
EVAL-X, then optimizing the metric scores is trivial.
At test time, for a given input,
we first use the prediction model 
to get $\hat y$.
Then, we can simply query the evaluator model
and output the encoded \emph{explanation} as any one input token
for which the evaluator model outputs $\hat y$. 
Even without access to the exact model weights of EVAL-X, 
we can train multiple approximations of EVAL-X with different random seeds, 
then use majority voting across the approximation models to output the final \emph{explanations} and predictions.

\subsection{Setup}
Here, we describe the setup for evaluating the efficacy of the two encoding methods 
for the EVAL-X metrics on the Movies dataset \citep{zaidan2008modeling, deyoung2019eraser}.
We use pre-trained BERT tokenizers and models for the prediction model 
and train it for 10 epochs with a batch size of 32 and a learning rate of 2e-5. 
We compute standard ACC and AUROC  
and the EVAL-X metric versions (i.e. eACC and eAUROC).
For the first encoding method, token-label correlation, 
we average the results over five random seeds of the evaluator model.
For the second encoding method, we train one evaluator model and four approximation models of different random seeds,
then use majority voting to combine the predictions and \emph{explanations}.

\subsection{Results}
\begin{figure*}[htb]
    \centering
        \includegraphics[width=.49\linewidth]{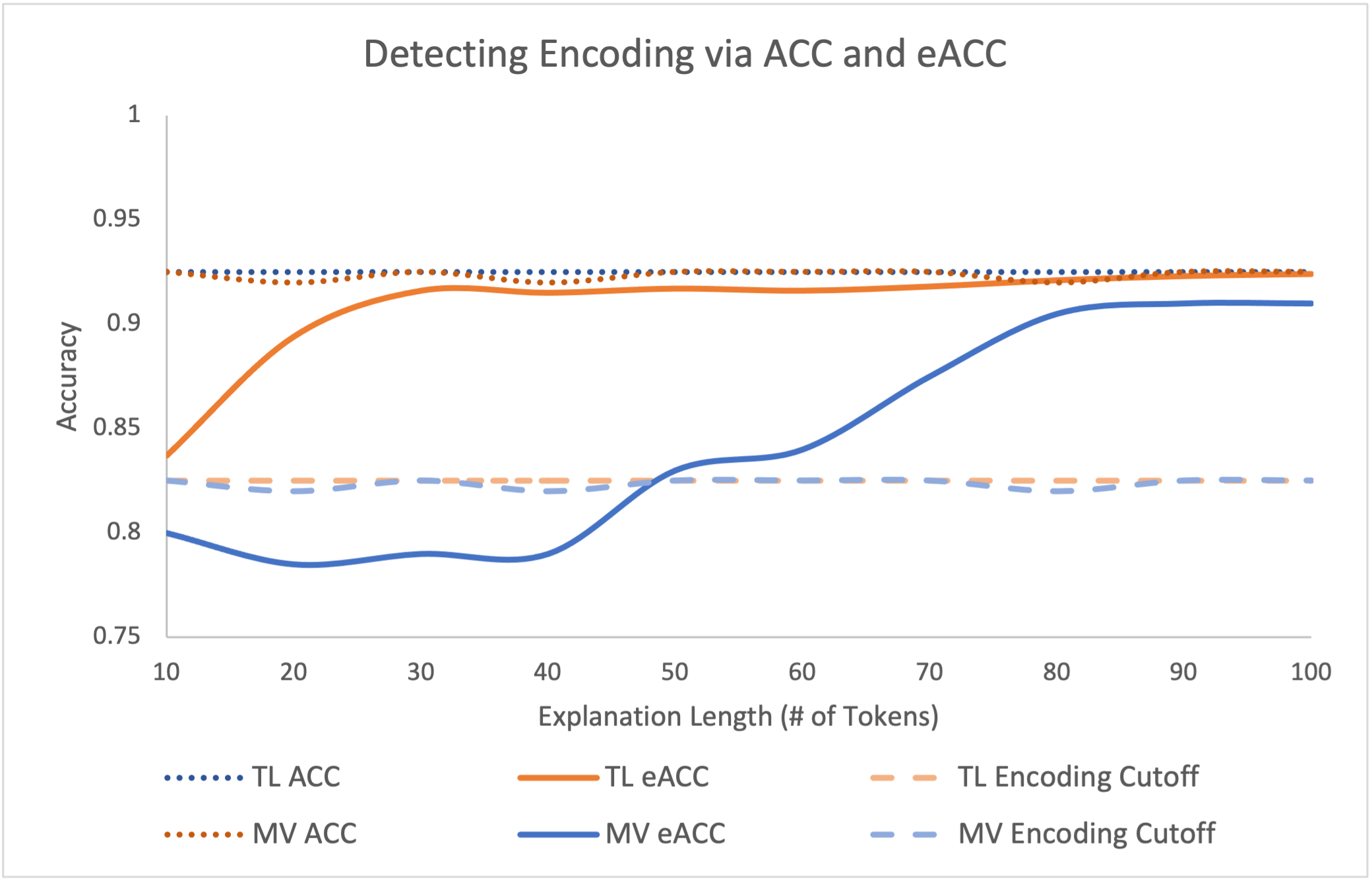}%
    \includegraphics[width=.49\linewidth]{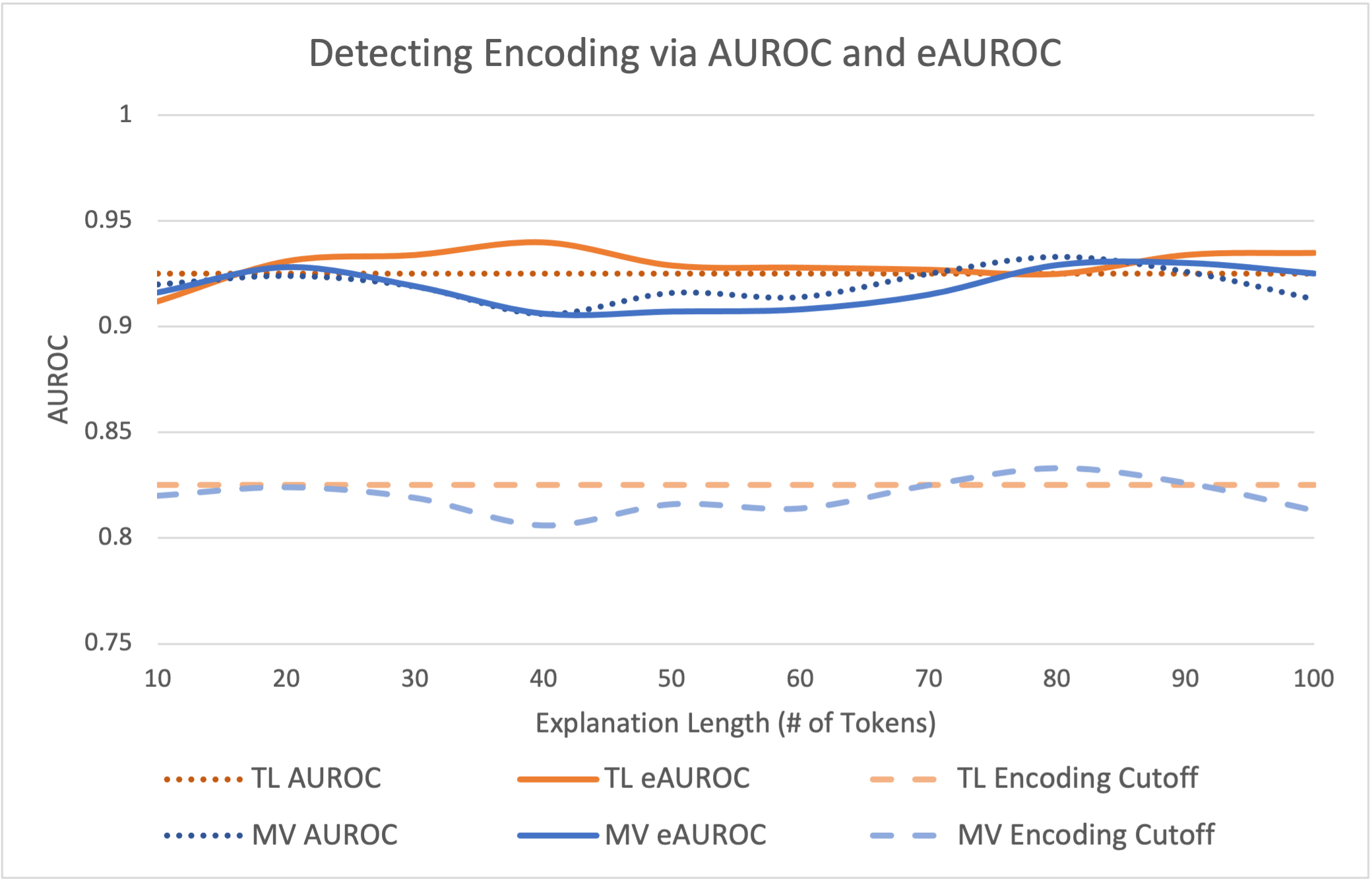} 
    
    \caption{
    These two figures evaluate whether our two encoding methods---\textcolor{orange}{Token-Label Likelihood Ratio (TL)} and 
    \textcolor{blue} {Majority-Voting (MV)}---are detected as encoding methods by the EVAL-X Metrics. We vary the length of the encoded \emph{explanations} along the x-axis.
    If a method's eACC and eAUORC are much lower (more than 10\% drop) than ACC and AUROC, 
    then it means that the method has been detected as an encoding method by the EVAL-X metrics. 
    We evaluate our eACC (left) and eAUROC (right)
    against their encoding cutoffs in dashed lines.
    From token lengths 10 to 100, 
    \textcolor{orange}{TL} \emph{explanations}' eACC and eAUROC are both above the encoding cutoff, 
    meaning \textcolor{orange}{TL} has not been detected as an encoding method. 
    For all token lengths, 
    \textcolor{blue}{MV} \emph{explanations}' eAUROC are above the cutoff, 
    while its eACC eventually increases to above the encoding cutoff starting from 50 tokens.}
    \label{fig:encoding}
\end{figure*}

We evaluate our two encoded saliency methods 
on the Movies dataset. 
Our methods achieve eACC and eAUROC 
above the encoding cutoff, meaning within a 10\% drop of the ACC and eAUROC.
This indicates that our methods have not been detected as encoded saliency methods by the EVAL-X metrics.

\paragraph{Method 1: Token-label Likelihood Ratio}
We encode the predictions into \emph{explanations} using 
token-label likelihood ratio.
The resulting eACC and eAUROC are both above the encoding cutoff of ACC and AUROC
across varying \emph{explanation} lengths from 10 to 100 (Fig. \ref{fig:encoding}).
On the Movies dataset,
with a length of 10 tokens,
our encoded \emph{explanations}' eACC is already above the encoding cutoff at a high of 83.7\%.
As we increase the encoded \emph{explanation} length,
eACC only increases till it matches ACC exactly at 92.5\%.
The success of this encoding method
shows how easy it is to score high on
the EVAL-X metrics
with encoded \emph{explanations} that are constructed
completely independently of interactions with the prediction model
(other than accessing the predicted labels on the original inputs).

\paragraph{Method 2: Evaluator Model Queries}
Using direct access to the evaluator model,
we can select any single token in a given input 
that results in the evaluator model predicting the label we wish to encode.
The resulting eACC and eAUROC would match ACC and AUROC exactly.
This contrasts directly with the metric's original motivation, where they claim a single feature, encoded \emph{explanation} could easily be detected as encoded.
Although it may be the case that a \emph{random} single input feature can be detected by their metric,
a single feature encoded by accessing the evaluator model can avoid being detected.

We then consider the scenario where we do not have direct access to the evaluator model.
In this case, we can train several approximations of the evaluator model.
This is possible since the training scheme is simple
and the data is the training set of our original prediction model.
The resulting, majority-voted \emph{explanations} achieve eACC and eAUROC above the encoding cutoff starting from a length of 50 tokens (Figure \ref{fig:encoding}).
These results demonstrate that it can be easy to trivially optimize for a metric that relies on an easily accessible or approximated evaluator model.

\section{Conclusion}

We have demonstrated that simple methods can achieve
substantially better and, sometimes, 
near-optimal scores on current metrics
for evaluating rationales 
\emph{without} producing rationales 
that anyone would reasonably claim as being more faithful.
While these metrics 
represent honest efforts to codify 
desiderata associated with such rationales,
we conclude that they are not suitable to function as benchmarks.
In general, few metrics capture all desiderata of interest. 
Accuracy does not capture all desiderata associated 
with image classification
and ROUGE score is a weak proxy for translation quality. 
However, for a quantitative metric
to function as a useful benchmark,
it must be the case that concerted efforts
to optimize this metric necessarily
bring about desired technological improvements. 
The effort to lower ImageNet error truly
required genuine advancements in computer vision
and efforts to increase ROUGE have revolutionized
machine translation. 
Efforts to optimize a metric, respecting the rules
of the game should not be regarded as mere ``gaming'';
inspiring such efforts is the very purpose of a benchmark. 
In general, when developing a metric, 
it often takes multiple iterations 
of proposals and criticisms to arrive at a useful formalism.
For example, in privacy, many formal notions of privacy
were proposed, and each in turn criticized, 
before the community arrived at the robust
and mathematically rigorous measure of differential privacy.
Likewise, many ways to quantify information 
were proposed before Shannon's seminal work. 
While the common word \emph{explanation}
may be hopelessly broad, 
we do not rule out the possibility
that measures might be proposed 
that rigorously capture some useful 
notion of \emph{saliency}. 
Our hope is that these results can inspire 
improved definitions capable 
of guiding methodological research.

\section*{Acknowledgements}
The authors gratefully acknowledge support from the NSF (FAI 2040929 and IIS2211955), UPMC, Highmark Health, Abridge, Ford Research, Mozilla, the PwC Center, Amazon AI, JP Morgan Chase, the Block Center, the Center for Machine Learning and Health, NSF CIF grant CCF1763734, the AI Research Institutes program supported by NSF and USDA-NIFA under award 2021-67021-35329, and the CMU Software Engineering Institute (SEI) via Department of Defense contract FA8702-15-D-0002.

\bibliographystyle{unsrtnat}
\bibliography{bibliography} 

\begin{thebibliography}{33}
\providecommand{\natexlab}[1]{#1}
\providecommand{\url}[1]{\texttt{#1}}
\expandafter\ifx\csname urlstyle\endcsname\relax
  \providecommand{\doi}[1]{doi: #1}\else
  \providecommand{\doi}{doi: \begingroup \urlstyle{rm}\Url}\fi

\bibitem[Lipton(2018)]{lipton2018mythos}
Zachary~C Lipton.
\newblock The mythos of model interpretability.
\newblock \emph{Communications of the ACM (CACM)}, 2018.

\bibitem[Pruthi et~al.(2022)Pruthi, Bansal, Dhingra, Soares, Collins, Lipton,
  Neubig, and Cohen]{pruthi2022evaluating}
Danish Pruthi, Rachit Bansal, Bhuwan Dhingra, Livio~Baldini Soares, Michael
  Collins, Zachary~C Lipton, Graham Neubig, and William~W Cohen.
\newblock Evaluating explanations: How much do explanations from the teacher
  aid students?
\newblock \emph{Transactions of the Association for Computational Linguistics},
  10:\penalty0 359--375, 2022.

\bibitem[Krishna et~al.(2022)Krishna, Han, Gu, Pombra, Jabbari, Wu, and
  Lakkaraju]{krishna2022disagreement}
Satyapriya Krishna, Tessa Han, Alex Gu, Javin Pombra, Shahin Jabbari, Steven
  Wu, and Himabindu Lakkaraju.
\newblock The disagreement problem in explainable machine learning: A
  practitioner's perspective.
\newblock \emph{arXiv preprint arXiv:2202.01602}, 2022.

\bibitem[Jethani et~al.(2021)Jethani, Sudarshan, Aphinyanaphongs, and
  Ranganath]{jethani2021have}
Neil Jethani, Mukund Sudarshan, Yindalon Aphinyanaphongs, and Rajesh Ranganath.
\newblock Have we learned to explain?: How interpretability methods can learn
  to encode predictions in their interpretations.
\newblock In \emph{International Conference on Artificial Intelligence and
  Statistics}, pages 1459--1467. PMLR, 2021.

\bibitem[Jacovi and Goldberg(2020)]{jacovi2020towards}
Alon Jacovi and Yoav Goldberg.
\newblock Towards faithfully interpretable nlp systems: How should we define
  and evaluate faithfulness?
\newblock \emph{arXiv preprint arXiv:2004.03685}, 2020.

\bibitem[DeYoung et~al.(2019)DeYoung, Jain, Rajani, Lehman, Xiong, Socher, and
  Wallace]{deyoung2019eraser}
Jay DeYoung, Sarthak Jain, Nazneen~Fatema Rajani, Eric Lehman, Caiming Xiong,
  Richard Socher, and Byron~C Wallace.
\newblock Eraser: A benchmark to evaluate rationalized nlp models.
\newblock \emph{arXiv preprint arXiv:1911.03429}, 2019.

\bibitem[Agarwal et~al.(2022)Agarwal, Krishna, Saxena, Pawelczyk, Johnson,
  Puri, Zitnik, and Lakkaraju]{agarwal2022openxai}
Chirag Agarwal, Satyapriya Krishna, Eshika Saxena, Martin Pawelczyk, Nari
  Johnson, Isha Puri, Marinka Zitnik, and Himabindu Lakkaraju.
\newblock Openxai: Towards a transparent evaluation of model explanations.
\newblock \emph{Advances in Neural Information Processing Systems},
  35:\penalty0 15784--15799, 2022.

\bibitem[Petsiuk et~al.(2018)Petsiuk, Das, and Saenko]{petsiuk2018rise}
Vitali Petsiuk, Abir Das, and Kate Saenko.
\newblock Rise: Randomized input sampling for explanation of black-box models.
\newblock \emph{arXiv preprint arXiv:1806.07421}, 2018.

\bibitem[Hooker et~al.(2019)Hooker, Erhan, Kindermans, and
  Kim]{hooker2019benchmark}
Sara Hooker, Dumitru Erhan, Pieter-Jan Kindermans, and Been Kim.
\newblock A benchmark for interpretability methods in deep neural networks.
\newblock \emph{Advances in neural information processing systems}, 32, 2019.

\bibitem[Serrano and Smith(2019)]{serrano2019attention}
Sofia Serrano and Noah~A Smith.
\newblock Is attention interpretable?
\newblock \emph{arXiv preprint arXiv:1906.03731}, 2019.

\bibitem[Covert et~al.(2021)Covert, Lundberg, and Lee]{covert2021explaining}
Ian~C Covert, Scott Lundberg, and Su-In Lee.
\newblock Explaining by removing: A unified framework for model explanation.
\newblock \emph{The Journal of Machine Learning Research}, 22\penalty0
  (1):\penalty0 9477--9566, 2021.

\bibitem[Samek et~al.(2015)Samek, Binder, Montavon, Bach, and
  Müller]{samek2015evaluating}
Wojciech Samek, Alexander Binder, Grégoire Montavon, Sebastian Bach, and
  Klaus-Robert Müller.
\newblock Evaluating the visualization of what a deep neural network has
  learned, 2015.

\bibitem[Nguyen(2018)]{nguyen2018comparing}
Dong Nguyen.
\newblock Comparing automatic and human evaluation of local explanations for
  text classification.
\newblock In \emph{Proceedings of the 2018 Conference of the North American
  Chapter of the Association for Computational Linguistics: Human Language
  Technologies, Volume 1 (Long Papers)}, pages 1069--1078, 2018.

\bibitem[Hase et~al.(2021)Hase, Xie, and Bansal]{hase2021out}
Peter Hase, Harry Xie, and Mohit Bansal.
\newblock The out-of-distribution problem in explainability and search methods
  for feature importance explanations.
\newblock \emph{Advances in neural information processing systems},
  34:\penalty0 3650--3666, 2021.

\bibitem[Ribeiro et~al.(2016)Ribeiro, Singh, and Guestrin]{ribeiro2016should}
Marco~Tulio Ribeiro, Sameer Singh, and Carlos Guestrin.
\newblock " why should i trust you?" explaining the predictions of any
  classifier.
\newblock In \emph{Proceedings of the 22nd ACM SIGKDD international conference
  on knowledge discovery and data mining}, pages 1135--1144, 2016.

\bibitem[Lundberg and Lee(2017)]{lundberg2017unified}
Scott~M Lundberg and Su-In Lee.
\newblock A unified approach to interpreting model predictions.
\newblock \emph{Advances in neural information processing systems}, 30, 2017.

\bibitem[Chan et~al.(2022)Chan, Kong, and Liang]{chan2022comparative}
Chun~Sik Chan, Huanqi Kong, and Guanqing Liang.
\newblock A comparative study of faithfulness metrics for model
  interpretability methods.
\newblock \emph{arXiv preprint arXiv:2204.05514}, 2022.

\bibitem[Slack et~al.(2020)Slack, Hilgard, Jia, Singh, and
  Lakkaraju]{slack2020fooling}
Dylan Slack, Sophie Hilgard, Emily Jia, Sameer Singh, and Himabindu Lakkaraju.
\newblock Fooling lime and shap: Adversarial attacks on post hoc explanation
  methods.
\newblock In \emph{Proceedings of the AAAI/ACM Conference on AI, Ethics, and
  Society}, pages 180--186, 2020.

\bibitem[Pruthi et~al.(2020)Pruthi, Gupta, Dhingra, Neubig, and
  Lipton]{pruthi19learning}
Danish Pruthi, Mansi Gupta, Bhuwan Dhingra, Graham Neubig, and Zachary~C.
  Lipton.
\newblock Learning to deceive with attention-based explanations.
\newblock In \emph{Annual Conference of the Association for Computational
  Linguistics (ACL)}, July 2020.

\bibitem[Wang et~al.(2020)Wang, Tuyls, Wallace, and Singh]{wang2020gradient}
Junlin Wang, Jens Tuyls, Eric Wallace, and Sameer Singh.
\newblock Gradient-based analysis of nlp models is manipulable.
\newblock \emph{arXiv preprint arXiv:2010.05419}, 2020.

\bibitem[Heo et~al.(2019)Heo, Joo, and Moon]{heo2019fooling}
Juyeon Heo, Sunghwan Joo, and Taesup Moon.
\newblock Fooling neural network interpretations via adversarial model
  manipulation.
\newblock In \emph{Advances in Neural Information Processing Systems
  (NeurIPS)}, 2019.

\bibitem[Dombrowski et~al.(2019)Dombrowski, Alber, Anders, Ackermann,
  M{\"u}ller, and Kessel]{dombrowski2019explanations}
Ann-Kathrin Dombrowski, Maximillian Alber, Christopher Anders, Marcel
  Ackermann, Klaus-Robert M{\"u}ller, and Pan Kessel.
\newblock Explanations can be manipulated and geometry is to blame.
\newblock \emph{Advances in neural information processing systems}, 32, 2019.

\bibitem[Ghorbani et~al.(2019)Ghorbani, Abid, and
  Zou]{ghorbani2019interpretation}
Amirata Ghorbani, Abubakar Abid, and James Zou.
\newblock Interpretation of neural networks is fragile.
\newblock In \emph{Proceedings of the AAAI conference on artificial
  intelligence}, volume~33, pages 3681--3688, 2019.

\bibitem[Anders et~al.(2020)Anders, Pasliev, Dombrowski, M{\"u}ller, and
  Kessel]{anders2020fairwashing}
Christopher Anders, Plamen Pasliev, Ann-Kathrin Dombrowski, Klaus-Robert
  M{\"u}ller, and Pan Kessel.
\newblock Fairwashing explanations with off-manifold detergent.
\newblock In \emph{International Conference on Machine Learning}, pages
  314--323. PMLR, 2020.

\bibitem[Zaidan and Eisner(2008)]{zaidan2008modeling}
Omar Zaidan and Jason Eisner.
\newblock Modeling annotators: A generative approach to learning from annotator
  rationales.
\newblock In \emph{Proceedings of the 2008 conference on Empirical methods in
  natural language processing}, pages 31--40, 2008.

\bibitem[Clark et~al.(2019)Clark, Lee, Chang, Kwiatkowski, Collins, and
  Toutanova]{clark2019boolq}
Christopher Clark, Kenton Lee, Ming-Wei Chang, Tom Kwiatkowski, Michael
  Collins, and Kristina Toutanova.
\newblock Boolq: Exploring the surprising difficulty of natural yes/no
  questions.
\newblock \emph{arXiv preprint arXiv:1905.10044}, 2019.

\bibitem[Lehman et~al.(2019)Lehman, DeYoung, Barzilay, and
  Wallace]{lehman2019inferring}
Eric Lehman, Jay DeYoung, Regina Barzilay, and Byron~C Wallace.
\newblock Inferring which medical treatments work from reports of clinical
  trials.
\newblock \emph{arXiv preprint arXiv:1904.01606}, 2019.

\bibitem[Thorne et~al.(2018)Thorne, Vlachos, Christodoulopoulos, and
  Mittal]{thorne2018fever}
James Thorne, Andreas Vlachos, Christos Christodoulopoulos, and Arpit Mittal.
\newblock Fever: a large-scale dataset for fact extraction and verification.
\newblock \emph{arXiv preprint arXiv:1803.05355}, 2018.

\bibitem[Khashabi et~al.(2018)Khashabi, Chaturvedi, Roth, Upadhyay, and
  Roth]{khashabi2018looking}
Daniel Khashabi, Snigdha Chaturvedi, Michael Roth, Shyam Upadhyay, and Dan
  Roth.
\newblock Looking beyond the surface: A challenge set for reading comprehension
  over multiple sentences.
\newblock In \emph{Proceedings of the 2018 Conference of the North American
  Chapter of the Association for Computational Linguistics: Human Language
  Technologies, Volume 1 (Long Papers)}, pages 252--262, 2018.

\bibitem[Devlin et~al.(2018)Devlin, Chang, Lee, and Toutanova]{devlin2018bert}
Jacob Devlin, Ming-Wei Chang, Kenton Lee, and Kristina Toutanova.
\newblock Bert: Pre-training of deep bidirectional transformers for language
  understanding.
\newblock \emph{arXiv preprint arXiv:1810.04805}, 2018.

\bibitem[Sundararajan et~al.(2017)Sundararajan, Taly, and
  Yan]{sundararajan2017axiomatic}
Mukund Sundararajan, Ankur Taly, and Qiqi Yan.
\newblock Axiomatic attribution for deep networks.
\newblock In \emph{International conference on machine learning}, pages
  3319--3328. PMLR, 2017.

\bibitem[Xu et~al.(2015)Xu, Ba, Kiros, Cho, Courville, Salakhudinov, Zemel, and
  Bengio]{xu2015show}
Kelvin Xu, Jimmy Ba, Ryan Kiros, Kyunghyun Cho, Aaron Courville, Ruslan
  Salakhudinov, Rich Zemel, and Yoshua Bengio.
\newblock Show, attend and tell: Neural image caption generation with visual
  attention.
\newblock In \emph{International conference on machine learning}, pages
  2048--2057. PMLR, 2015.

\bibitem[Treviso and Martins(2020)]{treviso2020explanation}
Marcos~V Treviso and Andr{\'e}~FT Martins.
\newblock The explanation game: Towards prediction explainability through
  sparse communication.
\newblock \emph{arXiv preprint arXiv:2004.13876}, 2020.

\end{thebibliography}
\appendix
\clearpage
\section{Additional Results for Sufficiency and Comprehensiveness}
\label{sec:eraser_appendix}
We show our faithfulness optimization results in Table~\ref{tab:eraser} and case detection accuracy in Table~\ref{tab:eraser_cd}  for datasets: Evidence Inference \citep{lehman2019inferring}, BoolQ \citep{clark2019boolq}, Movies \citep{zaidan2008modeling}, MultiRC \citep{khashabi2018looking}, and FEVER \citep{thorne2018fever}).

\begin{table*}[!htbp]
    \centering
    \caption{Gaming ERASER's Sufficiency and Comprehensiveness}
    \label{tab:eraser}
\begin{tabular}{lcccc}
\hline
             & \textbf{F1 Score} & \textbf{Comp.} & \textbf{Suff.} & \textbf{Comp.+Suff.}  \\

\hline 
\multicolumn{2}{l}{\textbf{Evidence   Inference}} &                &                 &                       \\
Attention              & 58.2              & 0.13           & -0.15           & -0.02                 \\
\rowcolor[HTML]{E7E6E6} 
Attention + meta-algo & 58.2              & 0.61           & -0.08           & 0.54                  \\
Gradient               & 58.3              & 0.15           & -0.12           & 0.04                  \\
\rowcolor[HTML]{E7E6E6} 
Gradient + meta-algo  & 58.3              & 0.61           & -0.10           & 0.51                  \\
LIME                   & 58.2              & 0.16           & -0.15           & 0.01                  \\
\rowcolor[HTML]{E7E6E6} 
LIME + meta-algo      & 58.2              & 0.66           & 0.14            & 0.79                  \\
Random                 & 58.2              & 0.05           & -0.21           & -0.16                 \\
\rowcolor[HTML]{E7E6E6} 
Random + meta-algo    & 58.2              & 0.65           & -0.15           & 0.50                  \\
\hline \textbf{BoolQ}                &                   &                &                 &                       \\
Attention              & 58.4              & 0.05           & -0.01           & 0.04                  \\
\rowcolor[HTML]{E7E6E6} 
Attention + meta-algo & 58.4              & 0.59           & 0.16            & 0.75                  \\
Gradient               & 58.4              & 0.03           & 0.00            & 0.04                  \\
\rowcolor[HTML]{E7E6E6} 
Gradient + meta-algo  & 58.4              & 0.73           & 0.25            & 0.98                  \\
LIME                   & 58.4              & 0.09           & 0.08            & 0.16                  \\
\rowcolor[HTML]{E7E6E6} 
LIME + meta-algo      & 58.4              & 0.73           & 0.26            & 1.00                  \\
Random                 & 58.4              & 0.01           & -0.06           & -0.05                 \\
\rowcolor[HTML]{E7E6E6} 
Random + meta-algo    & 58.4              & 0.65           & 0.12            & 0.77                  \\
\hline \textbf{Movies}               &                   &                &                 &                       \\
Attention              & 92.4              & 0.18           & -0.11           & 0.07                  \\
\rowcolor[HTML]{E7E6E6} 
Attention + meta-algo & 92.4              & 0.89           & -0.09           & 0.80                  \\
Gradient               & 92.4              & 0.26           & -0.08           & 0.18                  \\
\rowcolor[HTML]{E7E6E6} 
Gradient + meta-algo  & 92.4              & 0.83           & -0.09           & 0.74                  \\
LIME                   & 92.4              & 0.38           & -0.01           & 0.37                  \\
\rowcolor[HTML]{E7E6E6} 
LIME + meta-algo      & 92.4              & 0.82           & 0.00            & 0.82                  \\
Random                 & 92.4              & 0.01           & -0.06           & -0.05                 \\
\rowcolor[HTML]{E7E6E6} 
Random + meta-algo    & 92.4              & 0.65           & 0.12            & 0.77                  \\
\hline \textbf{MultiRC}              &                   &                &                 &                       \\
Attention              & 71.4              & 0.28           & -0.16           & 0.11                  \\
\rowcolor[HTML]{E7E6E6} 
Attention + meta-algo & 70.3              & 0.68           & -0.18           & 0.50                  \\
Gradient               & 71.4              & 0.26           & -0.23           & 0.04                  \\
\rowcolor[HTML]{E7E6E6} 
Gradient + meta-algo  & 70.7              & 0.68           & -0.20           & 0.48                  \\
LIME                   & 71.4              & 0.31           & -0.23           & 0.07                  \\
\rowcolor[HTML]{E7E6E6} 
LIME + meta-algo      & 71.0              & 0.77           & -0.04           & 0.73                  \\
Random                 & 71.4              & 0.10           & -0.39           & -0.29                 \\
\rowcolor[HTML]{E7E6E6} 
Random + meta-algo    & 71.4              & 0.75           & -0.29           & 0.47                  \\
\hline \textbf{FEVER}                &                   &                &                 &                       \\
Attention              & 90.7              & 0.13           & -0.15           & -0.02                 \\
\rowcolor[HTML]{E7E6E6} 
Attention + meta-algo & 90.7              & 0.61           & -0.08           & 0.54                  \\
Gradient               & 90.7              & 0.15           & -0.12           & 0.04                  \\
\rowcolor[HTML]{E7E6E6} 
Gradient + meta-algo  & 89.2              & 0.61           & -0.10           & 0.51                  \\
LIME                   & 90.7              & 0.09           & -0.23           & -0.14                 \\
\rowcolor[HTML]{E7E6E6} 
LIME + meta-algo      & 90.0              & 0.91           & -0.06           & 0.85                  \\
Random                 & 90.7              & 0.04           & -0.24           & -0.21                 \\
\rowcolor[HTML]{E7E6E6} 
Random + meta-algo    & 90.0              & 0.91           & -0.15           & 0.75   \\\hline           
\end{tabular}

\end{table*}

\begin{table*}[!htbp]
    \centering
    \caption{ERASER Case detector accuracy}
    \label{tab:eraser_cd}
    \begin{tabular}{lc}
    \hline
                                & Case detector Accuracy (\%) \\ \hline
    \textbf{Evidence Inference} &        \\ 
    Attention            & 78.6 \\
    Gradient             & 77.5 \\
    LIME                 & 88.9 \\
    Random               & 78.6 \\ \hline
    \textbf{BoolQ}              &  \\
    Attention            & 91.8 \\
    Gradient             & 99.3 \\
    LIME                 & 99.8 \\
    Random               & 92.2 \\ \hline
    \textbf{Movies}              &  \\
    Attention            & 93.3 \\
    Gradient             & 91.2 \\
    LIME                 & 93.7 \\
    Random               & 85.0 \\ \hline
    \textbf{MultiRC}              &  \\
    Attention            & 82.6 \\
    Gradient             & 81.7 \\
    LIME                 & 90.9 \\
    Random               & 82.3 \\ \hline
     \textbf{FEVER}              &       \\
     Attention            & 93.1  \\
     Gradient             & 91.6 \\
     LIME                 & 90.7 \\
     Random               & 91.5 \\ \hline
\end{tabular}
\end{table*}

\newpage
\section{Additional Results for EVAL-X Metrics}
We include the label recovery rate, ACC, AUROC, eACC, and eAUROC
for encoding method 1 (Token-label Likelihood Ratio) in Table \ref{tab:naive_bayes}
and for encoding method 2 (Majority Voting of Evaluator Model Approximations) in
Table \ref{tab:eval-meth1-movies}
on the Movies dataset \citep{zaidan2008modeling} in the ERASER benchmark \citep{deyoung2019eraser}.

For method 2 (Evaluator Model Queries), we compare using majority-voting of four evaluator model approximations to using only a single evaluator model approximation in Table \ref{tab:eval-meth1-movies} and Table \ref{tab:single-approx}.
We find that the EVAL-X scores are lower and have a higher variance when using a single approximation model. For the single evaluator model approximation experiments, we use one seed for the approximate model and four random seeds for the evaluator model.

\begin{table*}[!htbp]
\caption{EVAL-X Encoding Method 1: Naive Bayes Method}
    \centering
    \label{tab:naive_bayes}
    \begin{tabular}{cccccc}
    \toprule
    Num. of tokens & Label recovery rate (\%) & ACC (\%)   & eACC (\%)   & AUROC         & eAUROC        \\ 
    \midrule
1    & 100.0                   & 92.5 & 0.615$\pm$0.064   & 0.925 & 0.692$\pm$0.111      \\
5    & 100.0                   & 92.5 & 0.776$\pm$0.065   & 0.925 & 0.865$\pm$0.037     \\
10   & 100.0                   & 92.5 & 0.837$\pm$0.054    & 0.925 & 0.912$\pm$0.014      \\
20   & 100.0                   & 92.5 & 0.894$\pm$0.026   & 0.925 & 0.931$\pm$0.013     \\
50   & 100.0                   & 92.5 & 0.917$\pm$0.012   & 0.925 & 0.929$\pm$0.012     \\
100  & 100.0                   & 92.5 & 0.924$\pm$0.002    & 0.925 & 0.935$\pm$0.008   \\ \bottomrule
\end{tabular}
\end{table*}

\begin{table*}[htb]
    \centering
    \caption{EVAL-X Encoding Method: Majority Voting of Evaluator Model Approximations}
    \label{tab:eval-meth1-movies}
    \begin{tabular}{cccccc}
      \toprule
          Num. of tokens & Label recovery rate (\%) & ACC (\%)   & eACC (\%)   & AUROC (\%)        & eAUROC   (\%)     \\ 
      \midrule
      1 & 95.5 & 89.0 & 84.0 &  93.7 & 93.0\\
      
      10 & 100.0 & 92.5 & 80.0 & 92.0 & 91.6\\

      50 & 100.0 & 92.5 & 83.0 & 91.6 & 90.7\\

      70 & 100.0 & 92.5 & 87.5 & 92.5 & 91.5\\

      100 & 100.0 & 92.5 & 91.0 & 91.3 & 92.5\\
      \bottomrule
    \end{tabular}
\end{table*}

\begin{table*}[!htbp]
    \caption{EVAL-X Encoding Method: Single Evaluator Model Approximation}
    \centering
    \label{tab:single-approx}
    \begin{tabular}{cccccc}
    \toprule
    Num. of tokens & Label recovery rate (\%) & ACC (\%)   & eACC (\%)   & AUROC (\%)         & eAUROC (\%)       \\ 
    \midrule
    1             & 98.1 ± 2.4               & 90.9 ± 2.0 & 82.1 ± 11.0 & 90.9 ± 2.0 & 90.5 ± 2.5 \\
    5             & 99.1 ± 0.4               & 91.6 ±  0.4 & 80.9 ± 13.3 & 91.6 ± 0.4 & 87.4 ± 7.7 \\
    10            & 99.2 ± 0.6               & 91.7 ± 0.6  & 80.9 ± 13.3 & 91.7 ± 0.6 & 86.5 ± 7.6 \\
    50            & 98.7 ±1.3                & 91.5 ± 1.5 & 83.3 ± 10.8 & 91.5 ± 1.5 & 90.1 ± 4.7 \\
    70            & 99.2 ± 0.8                & 92.0 ± 0.5  & 83.1 ± 10.7 & 92.9 ± 0.5  & 91.0 ± 3.5  \\
    100           & 98.5 ± 2.1               & 91.3 ± 1.6 & 83.4 ± 10.1 & 91.2 ± 1.6 & 91.3 ± 3.6 \\ \bottomrule
    \end{tabular}
\end{table*}

\end{document}